\documentclass[letterpaper]{article} 
\usepackage[preprint]{aaai2027}  
\usepackage[hyphens]{url}  
\usepackage{graphicx} 
\usepackage{natbib}  
\usepackage{caption} 
\usepackage{algorithm}
\usepackage{algorithmic}
\usepackage{amsfonts}
\usepackage{amsmath}
\usepackage{newfloat}
\usepackage{listings}
\DeclareCaptionStyle{ruled}{labelfont=normalfont,labelsep=colon,strut=off} 
\floatstyle{ruled}
\newfloat{listing}{tb}{lst}{}
\floatname{listing}{Listing}

\usepackage{booktabs}

\title{Interpretable Orthogonal Latent Diffusion with
Scale-Selective Uncertainty for Medical Image Super-Resolution}

\author{
    Onkar Jadhav\textsuperscript{\rm 1}\corresponding,
    Tim French\textsuperscript{\rm 2}
}

\affiliations{
    \textsuperscript{\rm 1}School of Earth and Oceans, 
    UWA Oceans Institute\\
    \textsuperscript{\rm 2}School of Physics, Mathematics and Computing, 
    Computer Science and Software Engineering\\
    The University of Western Australia, Crawley, WA, Australia\\
    onkar.jadhav.ai@gmail.com, tim.french@uwa.edu.au
}

\begin{document}

\maketitle
\begin{abstract}
Diffusion models are increasingly used for medical image reconstruction and super-resolution, but their samples may contain plausible fine-scale hallucinations that are difficult to identify from the reconstructed image alone. We propose Med-PODiff, a patch-wise proper orthogonal decomposition (POD) diffusion framework for brain MRI super-resolution that performs generative reconstruction in compact and interpretable local coefficient spaces. Unlike pixel-space or learned-latent diffusion models, Med-PODiff uses the POD structure to separate low-rank anatomical variation from fine-detail residual variation, enabling predictive uncertainty to be decomposed into total, coarse, and fine components. Building on this decomposition, we define a pixel-wise hallucination-risk score, $H_p$, that emphasizes regions where fine-detail uncertainty dominates coarse structural uncertainty. On brain MRI, Med-PODiff outperforms representative diffusion baselines in reconstruction quality while producing well-calibrated predictive intervals, with total uncertainty detecting high-error regions at AUROC 0.954. More importantly, the decomposed uncertainty is scale-selective: coarse and fine uncertainty preferentially predict errors at the corresponding scales, with $H_p$ localizing fine error at AUROC $0.906$. Controlled injections and retrospective Cartesian $k$-space undersampling further support fine-detail localization and degradation robustness. The discarded POD tail is only $3.8\%$ of the retained fine-error magnitude, and Med-PODiff improves downstream segmentation consistency while requiring substantially less inference time than pixel-space diffusion.
\end{abstract}
\section{Introduction}
Diffusion models can generate sharp and anatomically plausible medical image reconstructions while representing the ambiguity inherent in super-resolution. However, their generative flexibility may also introduce plausible details that are insufficiently supported by the input. Such hallucination-prone errors are particularly concerning in medical imaging because false boundaries or tissue patterns may appear credible to clinicians and downstream algorithms.

Predictive uncertainty can identify unreliable reconstruction regions, and Bayesian, diffusion-based, and stochastic medical reconstruction methods have produced pixel-wise uncertainty maps that correlate with reconstruction error~\citep{tanno2017bayesian,luo2023bayesian,narnhofer2022bayesian,huang2025enhancing,giannakopoulos2026pixelwise}. These methods nevertheless represent uncertainty primarily as a single spatial field. Consequently, broad structural ambiguity and localized fine-detail uncertainty are conflated, despite their different implications for reconstruction reliability.

Related methods introduce scale through high-frequency diffusion, wavelets, or multiscale priors~\citep{cao2024high,huang2025enhancing,chand2024multi}, but do not decompose sample-based reconstruction uncertainty into separately validated coarse and fine components. Hallucination assessment instead uses measurement/null-space decomposition, distributional indices, or reference-based frequency analysis~\citep{bhadra2021hallucinations,tivnan2024hallucination,kc2026sfrc}; these approaches require measurements, distributions, or references rather than deriving an inference-time risk map directly from predictive uncertainty. Posterior-principal-direction methods model correlated restoration uncertainty~\citep{nehme2023nppc,belhasin2023volume}, while segmentation-guided MRI sampling probes downstream ambiguity~\citep{morshuis2024segmentation}; neither yields the energy-ordered coarse--fine uncertainty decomposition studied here.

Building on PODiff~\citep{jadhav2026podiff} and its patch-wise
extension, Patch-PODiff-ViT~\citep{patchPODiff}, we
introduce Med-PODiff for brain MRI super-resolution. While
these frameworks establish structured latent diffusion and
analytic uncertainty propagation, Med-PODiff exploits the
orthogonal latent geometry to decompose predictive uncertainty
into scale-selective components and identify fine-detail
hallucination risk.

Med-PODiff represents non-overlapping image patches in compact, energy-ordered POD coefficient spaces and performs conditional diffusion over the resulting token sequence. It partitions the retained POD modes into dominant low-order and remaining high-order groups. Uncertainty estimated from the generated coefficient ensemble is propagated exactly through the fixed linear decoder to obtain total, coarse, and fine spatial uncertainty maps. Since POD modes are ordered by explained variance rather than spatial frequency, these labels are operational and their scale selectivity is validated empirically.

We further introduce $H_p$, a continuous pixel-wise risk score that emphasizes locations where fine uncertainty is both large and dominant relative to coarse uncertainty. It requires no high-resolution reference at inference and is interpreted as a ranking score for fine-detail reconstruction risk, not as a calibrated hallucination probability. To our knowledge, Med-PODiff is the first medical image reconstruction framework to decompose sample-based predictive uncertainty over an energy-ordered orthogonal basis into separately validated structural and detail components and use their relative dominance to localize fine-detail reconstruction errors and downstream segmentation disagreements.

Our contributions are: (i) an exactly propagated POD-structured decomposition of predictive uncertainty into total, coarse, and fine components; (ii) the reference-free $H_p$ score for localizing fine-detail reconstruction risk; (iii) empirical validation through same- and cross-scale error
localization, controlled detail injection, cutoff and $k$-space
robustness, calibration, and POD-tail analysis; and (iv) evidence that the resulting uncertainty representation supports downstream segmentation-disagreement localization while retaining the efficiency of structured latent diffusion.
\section{Method}
\label{sec:method}

\paragraph{Overview.}
Med-PODiff adopts the structured latent diffusion formulation
of Patch-PODiff-ViT~\citep{patchPODiff} and extends it
to obtain scale-specific uncertainty and hallucination-risk
maps for medical image reconstruction.
A high-resolution image is partitioned into non-overlapping patches, projected onto a fixed training-derived POD basis, and represented as standardized coefficient tokens, over which conditional diffusion is performed using a Vision Transformer. The fixed linear decoder enables exact propagation of coefficient uncertainty through the retained basis and its separation into low- and high-order components. The principal contributions are this uncertainty decomposition, its scale-selective interpretation, and the resulting hallucination-risk formulation.

\paragraph{Patch-wise POD representation.}
Let $x^{\mathrm{HR}}\in\mathbb R^{H\times W}$ denote a high-resolution target, partitioned into $P$ non-overlapping $p\times p$ patches $\mathbf u_p\in\mathbb R^s$, where $s=p^2$. A single shared POD basis \citep{sirovich1987turbulence} is constructed from all training patches. Let $\bar{\mathbf u}$ and $\mathbf C_u$ denote their mean and empirical covariance. Its eigenpairs satisfy
\begin{equation}
\mathbf C_u\boldsymbol\phi_k=\lambda_k\boldsymbol\phi_k,
\qquad
\lambda_1\geq\cdots\geq\lambda_s\geq0,
\end{equation}
and we retain
\begin{equation}
K=\min\left\{k:
\frac{\sum_{j=1}^{k}\lambda_j}
     {\sum_{j=1}^{s}\lambda_j}
\geq\eta\right\},
\qquad \eta=0.99.
\label{eq:energy}
\end{equation}
With $\mathbf\Phi=[\boldsymbol\phi_1,\ldots,\boldsymbol\phi_K]$ and $\mathbf D=\operatorname{diag}(\sqrt{\lambda_1},\ldots,\sqrt{\lambda_K})$, each patch is encoded, standardized, and reconstructed as
\begin{equation}
\mathbf a_p=\mathbf\Phi^\top(\mathbf u_p-\bar{\mathbf u}),
\qquad
\mathbf z_{0,p}=\mathbf D^{-1}\mathbf a_p,
\qquad
\mathbf u_p^{(K)}=\bar{\mathbf u}+\mathbf\Phi\mathbf a_p .
\label{eq:pod_encoding}
\end{equation}
The row-stacked standardized coefficients form the clean token sequence $\mathbf Z_0=[\mathbf z_{0,1},\ldots,\mathbf z_{0,P}]^\top\in\mathbb R^{P\times K}$. Both the POD encoder and decoder remain fixed throughout diffusion training.

\paragraph{Conditional diffusion in coefficient space.}
The low-resolution observation is bicubically upsampled to the target grid, $x^{\mathrm{up}}=\mathcal U(x^{\mathrm{LR}})$, and partitioned using the same patch layout. For upsampled patch $\mathbf v_p$, the aligned conditioning token is
\begin{equation}
\mathbf c_p
=\mathbf D^{-1}\mathbf\Phi^\top(\mathbf v_p-\bar{\mathbf u}),
\end{equation}
and row-stacking produces $\mathbf C\in\mathbb R^{P\times K}$. The standardized target coefficients are diffused \citep{ho2020denoising} according to
\begin{equation}
\mathbf Z_t
=\sqrt{\bar{\alpha}_t}\mathbf Z_0
+\sqrt{1-\bar{\alpha}_t}\boldsymbol\epsilon,
\qquad
\boldsymbol\epsilon\sim\mathcal N(\mathbf0,\mathbf I),
\label{eq:forward_diffusion}
\end{equation}
where $\bar{\alpha}_t=\prod_{j=1}^{t}\alpha_j$ follows a cosine schedule \citep{nichol2021improved} over $T=1000$ timesteps. The denoiser is trained to predict the injected noise:
\begin{equation}
\mathcal L_{\mathrm{diff}}
=
\mathbb E_{\mathbf Z_0,t,\boldsymbol\epsilon}
\left[
\left\|
\boldsymbol\epsilon-
\boldsymbol\epsilon_\theta(\mathbf Z_t,\mathbf C,t)
\right\|_2^2
\right],
\quad
t\sim\mathcal U\{1,\ldots,T\}.
\label{eq:diffusion_loss}
\end{equation}
Noisy and conditioning tokens are projected separately and fused at corresponding patch locations:
\begin{equation}
\mathbf h_p
=
\mathbf W_{\mathrm{in}}\mathbf z_{t,p}
+\mathbf W_{\mathrm{cond}}\mathbf c_p
+\mathbf e_p,
\label{eq:conditioning}
\end{equation}
where $\mathbf e_p$ is a two-dimensional positional embedding. The resulting sequence is processed by a Vision Transformer \citep{dosovitskiy2020image} comprising multi-head self-attention, MLP blocks, and timestep-conditioned Adaptive LayerNorm. A linear output head predicts noise in $\mathbb R^{P\times K}$. Thus, conditioning remains aligned patch by patch, while self-attention captures dependencies between spatially separated patches.

\paragraph{Reconstruction and predictive uncertainty.}
At inference, we generate $M$ independent coefficient samples using DDIM \citep{song2020denoising}. Each sample is de-standardized and decoded as
\begin{equation}
\hat{\mathbf a}_p^{(m)}
=\mathbf D\hat{\mathbf z}_p^{(m)},
\qquad
\hat{\mathbf u}_p^{(m)}
=\bar{\mathbf u}+\mathbf\Phi\hat{\mathbf a}_p^{(m)}.
\label{eq:decoding}
\end{equation}
Since patches do not overlap, $\hat{x}^{(m)}$ is obtained by direct tiling. Its predictive mean and total uncertainty are
\begin{align}
\bar{x}(\mathbf r)
&=\frac{1}{M}\sum_{m=1}^{M}\hat{x}^{(m)}(\mathbf r),\\
U_{\mathrm{tot}}(\mathbf r)
&=\left[
\frac{1}{M-1}\sum_{m=1}^{M}
\bigl(\hat{x}^{(m)}(\mathbf r)-\bar{x}(\mathbf r)\bigr)^2
\right]^{1/2}.
\label{eq:total_uncertainty}
\end{align}
Equivalently, if $\widehat{\mathbf\Sigma}_{a,p}$ is the empirical covariance of the de-standardized coefficient samples, the decoded patch covariance is
\begin{equation}
\widehat{\mathbf\Sigma}_{u,p}
=\mathbf\Phi\widehat{\mathbf\Sigma}_{a,p}\mathbf\Phi^\top.
\label{eq:covariance_propagation}
\end{equation}
Its diagonal is exactly the sample variance of the decoded patch. Since every pixel belongs to one non-overlapping patch, cross-patch covariances are unnecessary for the marginal pixel-wise variance map; they would be required only for the complete image covariance.

\paragraph{Low- and high-order uncertainty decomposition.}
We partition the retained basis at $k_c$:
\begin{equation}
\mathbf\Phi_c=[\boldsymbol\phi_1,\ldots,\boldsymbol\phi_{k_c}],
\qquad
\mathbf\Phi_f=[\boldsymbol\phi_{k_c+1},\ldots,\boldsymbol\phi_K].
\label{eq:basis_split}
\end{equation}
For sample $m$, the corresponding components are
\begin{align}
\hat{\mathbf u}_{p,c}^{(m)}
&=\bar{\mathbf u}
+\mathbf\Phi_c\hat{\mathbf a}_{p,c}^{(m)},\\
\hat{\mathbf u}_{p,f}^{(m)}
&=\mathbf\Phi_f\hat{\mathbf a}_{p,f}^{(m)},
\qquad
\hat{\mathbf u}_p^{(m)}
=\hat{\mathbf u}_{p,c}^{(m)}
+\hat{\mathbf u}_{p,f}^{(m)}.
\label{eq:component_decoding}
\end{align}
Tiling gives $\hat{x}_c^{(m)}$ and $\hat{x}_f^{(m)}$, whose sample standard deviations define $U_c$ and $U_f$ using Eq.~\eqref{eq:total_uncertainty}. Because the coefficient covariance may contain cross-group correlations,
\begin{equation}
U_{\mathrm{tot}}^2
=U_c^2+U_f^2+
2\,\operatorname{Cov}(\hat{x}_c,\hat{x}_f),
\label{eq:cross_covariance}
\end{equation}
and we do not impose variance additivity. POD modes are ordered by explained training variance rather than spatial frequency; therefore, \emph{coarse} and \emph{fine} are operational labels whose spatial selectivity is evaluated in the scale-selectivity experiments. The fine group contains the remaining retained modes $k_c+1,\ldots,K$, whereas modes above $K$ constitute the discarded truncation tail.

\paragraph{Scale-specific errors.}
The target is decomposed using the same retained basis. Let $x_c^{\mathrm{HR}}$ and $x_f^{\mathrm{HR}}$ denote the target components reconstructed from modes $1,\ldots,k_c$ and $k_c+1,\ldots,K$, respectively. We define
\begin{equation}
e_{\mathrm{tot}}=|x^{\mathrm{HR}}-\bar{x}|,
\qquad
e_c=|x_c^{\mathrm{HR}}-\bar{x}_c|,
\qquad
e_f=|x_f^{\mathrm{HR}}-\bar{x}_f|.
\label{eq:component_errors}
\end{equation}
Thus, $e_f$ measures error within the retained high-order POD subspace and excludes the discarded truncation tail.

\paragraph{Pixel-wise hallucination-risk score.}
Fine uncertainty is most concerning when it is both large and dominant relative to the low-order component. We therefore define
\begin{equation}
H_p(\mathbf r)
=
U_f(\mathbf r)
\frac{U_f(\mathbf r)}
{U_c(\mathbf r)+U_f(\mathbf r)+\epsilon},
\qquad
\epsilon=10^{-8}.
\label{eq:hp}
\end{equation}
The ratio measures the relative dominance of fine uncertainty, while the leading $U_f$ suppresses large ratios where both components are negligible. Consequently, $0\leq H_p(\mathbf r)\leq U_f(\mathbf r)$, and $H_p$ retains the units of predictive uncertainty. It is used as a continuous ranking score for fine-detail reconstruction risk, not as a calibrated hallucination probability or binary mask.
\section{Experimental Setup}
\label{sec:experiments}

\subsection{Datasets}
\label{sec:datasets}
We evaluate Med-PODiff on $4\times$ brain MRI super-resolution using the
multicoil brain subset of fastMRI~\citep{zbontar2018fastmri}: training
batches 0--2 (1,347 volumes; 19,090 slices), validation batch 0 (460
volumes; 1,808 slices), and test batch 0 (186 volumes; 2,130 slices).
All slices are converted to magnitude images, center-cropped and resized
to $256\times256$, and normalized to $[0,1]$. Low-resolution inputs are
produced by $4\times$ bicubic downsampling followed by bicubic upsampling
to the target grid, giving spatially aligned conditioning inputs. The
same preprocessing is used for all methods. Unless otherwise stated,
results are computed on the full $2{,}130$-image test set; the
scale-specific analysis uses a fixed random subset of 200 test images.

\subsection{Baselines and Implementation Details}
\label{sec:baselines}

We compare Med-PODiff against five baselines spanning deterministic, pixel-space, learned-latent, transformer, and non-patchwise POD approaches.

\textbf{Bicubic:} A deterministic, non-learned baseline that upsamples the low-resolution input directly to the target resolution via bicubic interpolation, with no conditioning or learned parameters.

\textbf{PixelDiff:} A pixel-space diffusion model using a convolutional U-Net denoiser with a symmetric encoder--decoder, four resolution levels, base width $C=128$, channel widths $(C,2C,4C,8C)$, and skip connections. It is trained with a DDPM objective and conditioned on the bicubically upsampled low-resolution input via channel concatenation, following the residual ($x^{\mathrm{HR}}-x^{\mathrm{up}}$) diffusion paradigm of CorrDiff~\citep{mardani2025residual}. 

\textbf{VAE-LDM:} A learned-latent diffusion model
\citep{rombach2022high}. A convolutional VAE with base width $C=128$
compresses $256\times256$ inputs to a $64\times64\times4$ latent
(selected on validation reconstruction quality); a diffusion model is
then trained in this latent space using the same U-Net used for PixelDiff.

\textbf{DiT:} A Diffusion Transformer~\citep{peebles2023dit} operating directly on $16\times16$ pixel patches without POD compression. It uses the same Vision Transformer configuration as Med-PODiff ($d_{\mathrm{model}}=512$, $L=12$ layers, $H=8$ attention heads) and the same patch sequence length, isolating the effect of POD compression from denoiser capacity.

\textbf{Full-field POD diffusion:} A non-patchwise POD baseline that uses a single global POD basis over the full image rather than a patchwise basis, with the retained mode count $K$ selected by the same $\eta=0.99$ energy criterion as Med-PODiff. This isolates the contribution of patchwise, as opposed to global, POD structure.

\textbf{Med-PODiff Implementation Details:} Non-overlapping $16\times16$ patches (stride 16) are represented using
a shared training-only POD basis retaining $K=128$ modes at
$\eta=0.99$; the resulting encoder and decoder remain fixed during
diffusion training. The denoiser is a 12-block Vision Transformer
($d_{\mathrm{model}}=512$, $H=8$, MLP ratio $4.0$; approximately 40M
parameters). It is trained for 500K steps with batch size 64 using
AdamW (learning rate and weight decay $10^{-4}$), gradient clipping at
1.0, and EMA decay 0.9995. We use a cosine schedule with $T=1{,}000$
training timesteps and DDIM with $S=100$ inference steps; unless
otherwise stated, uncertainty uses $M=100$ samples. Each reported result is obtained from 5 runs. All baselines use
the same optimizer, learning rate, and preprocessing, with
hyperparameters selected on validation and fixed for testing.
All models are trained on AMD Instinct MI250X GPUs.

\subsection{Evaluation Metrics}
\label{sec:metrics}
Reconstruction quality is measured using RMSE, MAE, PSNR, and
SSIM~\citep{wang2004image} between the predictive mean and high-resolution
target. Probabilistic performance uses CRPS~\citep{gneiting2007strictly},
four-level mean absolute calibration error ($\mathrm{MACE}_4$), and empirical
coverage at $50\%$, $70\%$, $90\%$, and $95\%$.
For error localization we define
$e_i(\mathbf r)=|\bar{x}_i(\mathbf r)-x_i^{\mathrm{HR}}(\mathbf r)|$,
restrict evaluation to the reference foreground
$\Omega_i^{\mathrm{HR}}=\{\mathbf r:x_i^{\mathrm{HR}}(\mathbf r)>0\}$, and
take positives to be the top $10\%$ of foreground pixels by error per image.
Scores ($U_{\mathrm{tot}}$, $U_c$, $U_f$, $H_p$) are evaluated against total,
coarse, and fine errors using AUROC, AUPRC, precision among the top $10\%$
highest-scoring pixels (P@10), and Pearson $r$, computed per image and
averaged across images.
For injected-detail localization we additionally report
sFRC~\citep{kc2026sfrc}, a reference-based comparator with continuous patch
score $1-\nu_c/\nu_{\mathrm{Nyq}}$ averaged pixelwise over overlapping
patches; its patch size and FRC threshold are selected on validation and
fixed for test.
\paragraph{Downstream segmentation.}
We apply the fixed SynthSeg model~\cite{billot2023synthseg} distributed with FreeSurfer 8.0.0 to each high-resolution and reconstructed three-dimensional volume using robust parcellation, volume, and quality-control modes, without fine-tuning on fastMRI. Since manual labels are unavailable, the high-resolution-derived segmentation serves as the reference. For a SynthSeg label map $S(x)$, we define binary foreground support as $F(x)(\mathbf r)=\mathbb I[S(x)(\mathbf r)>0]$. We report foreground Dice, IoU, symmetric HD95 in voxel units, and relative volume error $\left||F(\hat{x})|-|F(x^{\mathrm{HR}})|\right|/|F(x^{\mathrm{HR}})|$. Metrics are computed per volume and summarized as mean $\pm$ standard deviation across the matched test volumes.

\paragraph{Segmentation-disagreement localization.}
For slice $i$, let $F_i^{\mathrm{rec}}=F(\bar{x}_i)$ and $F_i^{\mathrm{HR}}=F(x_i^{\mathrm{HR}})$. We define foreground disagreement as
\begin{equation}
D_i(\mathbf r)=\mathbb I\!\left[F_i^{\mathrm{rec}}(\mathbf r)\neq F_i^{\mathrm{HR}}(\mathbf r)\right].
\label{eq:seg_disagreement}
\end{equation}
All scores are evaluated within the union foreground $\Omega_i=\{\mathbf r:F_i^{\mathrm{rec}}(\mathbf r)=1\ \lor\ F_i^{\mathrm{HR}}(\mathbf r)=1\}$. Total uncertainty, fine uncertainty, and $H_p$ are evaluated as predictors of $D_i$. AUROC, AUPRC, and P@10 are computed pixelwise for each valid slice and then averaged. Pearson $r$ instead measures the slice-level association between the mean score over $\Omega_i$ and the disagreement rate $d_i=|\Omega_i|^{-1}\sum_{\mathbf r\in\Omega_i}D_i(\mathbf r)$.
\section{Results}
\label{sec:results}

\subsection{Reconstruction Quality}
\label{sec:reconstruction_quality}
Table~\ref{tab:reconstruction_quality} shows that Med-PODiff achieves the best observed mean performance across all four metrics. Relative to DiT, it reduces RMSE by $9.9\%$ and MAE by $7.2\%$,
while improving PSNR by $0.905$\,dB and SSIM by $0.0031$. Thus, the structured POD representation improves reconstruction fidelity while additionally enabling the uncertainty decomposition studied below. All metrics are mean over 5 runs; per-run $SD<0.001$ for RMSE/MAE and $< 0.002$ for SSIM for all methods. 
\begin{table}[t]
\centering
\small
\setlength{\tabcolsep}{4.2pt}
\begin{tabular}{lcccc}
\hline
Method & RMSE $\downarrow$ & MAE $\downarrow$ &
PSNR $\uparrow$ & SSIM $\uparrow$ \\
\hline
Bicubic & 0.08980 & 0.05960 & 20.934 & 0.7508 \\
PixelDiff & 0.05201 & 0.03452 & 25.678 & 0.8988 \\
VAE-LDM & 0.05420 & 0.03597 & 25.320 & 0.8684 \\
DiT & 0.05120 & 0.03398 & 25.815 & 0.9089 \\
Full-field POD & 0.05242 & 0.03479 & 25.610 & 0.8696 \\
\textbf{Med-PODiff} &
\textbf{0.04611} & \textbf{0.03155} &
\textbf{26.720} & \textbf{0.9120} \\
\hline
\end{tabular}
\caption{Brain MRI $4\times$ super-resolution quality on the full test
set ($N=2{,}130$ slices). Metrics are computed between the predictive
mean and the high-resolution target.}
\label{tab:reconstruction_quality}
\end{table}
\subsection{Uncertainty Calibration}
\label{sec:calibration}

Table~\ref{tab:calibration} evaluates CRPS, four-level mean absolute calibration error ($\mathrm{MACE}_4$), and empirical coverage. Med-PODiff achieves the best CRPS ($0.0239$) and $\mathrm{MACE}_4$ ($0.0107$), improving over DiT by $4.8\%$ and $45.7\%$, respectively. It is also closest to nominal coverage at all reported levels, with only mild undercoverage at $50\%$. Unlike DiT, which shifts from undercoverage at $50\%$--$70\%$ to overcoverage at $90\%$--$95\%$, Med-PODiff remains well calibrated without relying on uniformly wide predictive intervals.

\begin{table}[t]
\centering
\small
\setlength{\tabcolsep}{1.4pt}
\renewcommand{\arraystretch}{0.98}
\begin{tabular}{@{}lcccccc@{}}
\hline
& & & \multicolumn{4}{c}{Empirical coverage (\%)} \\
\cline{4-7}
Method & CRPS$\downarrow$ & MACE$_4\downarrow$
& 50 & 70 & 90 & 95 \\
\hline
PixelDiff
& 0.0260 & 0.0471 & 44.42 & 64.36 & 86.18 & 91.22 \\
VAE-LDM
& 0.0300 & 0.0711 & 40.68 & 58.96 & 86.11 & 90.80 \\
DiT
& 0.0251 & 0.0197 & 46.90 & 67.96 & 91.59 & 96.14 \\
Full-field POD
& 0.0258 & 0.0367 & 45.05 & 65.28 & 87.54 & 92.47 \\
\textbf{Med-PODiff}
& \textbf{0.0239} & \textbf{0.0107}
& \textbf{47.58} & \textbf{68.95}
& \textbf{90.32} & \textbf{94.50} \\
\hline
\end{tabular}
\caption{Uncertainty calibration on the full test set
($N=2{,}130$ slices), using $M=100$ samples per input.
Coverage entries are percentages; $\mathrm{MACE}_4$ averages the
absolute coverage deviation at the four nominal levels. Full-field POD
denotes Full-field POD diffusion.}
\label{tab:calibration}
\end{table}
\subsection{Detection of High-Error Regions}
\label{sec:hallucination_detection}

We define high-error locations as the top $10\%$ of foreground
pixels ranked by absolute reconstruction error and evaluate how well
predictive uncertainty identifies them over the full test set
(Table~\ref{tab:hallucination_detection}). Med-PODiff total uncertainty
performs best, with AUROC $0.954$, AUPRC $0.768$, and P@10 $0.610$,
improving over DiT by $0.051$, $0.087$, and $0.023$, respectively.
Fine uncertainty nearly matches total uncertainty
(AUROC $0.951$, AUPRC $0.759$, P@10 $0.600$), while coarse uncertainty
remains informative but weaker (AUROC $0.926$). This suggests that much
of the spatial error signal lies in the higher retained POD modes, but
does not by itself establish scale selectivity; we test this next using
matched- and cross-scale uncertainty--error pairs.

\begin{table}[t]
\centering
\small
\setlength{\tabcolsep}{2.5pt}
\renewcommand{\arraystretch}{0.95}
\begin{tabular}{lcccc}
\hline
Risk score & AUROC$\uparrow$ & AUPRC$\uparrow$ &
P@10$\uparrow$ \\
\hline
Med-PODiff (Total UQ) &
\textbf{0.954} & \textbf{0.768} &
\textbf{0.610} \\
Med-PODiff (Fine UQ) & 0.951 & 0.759 & 0.600  \\
Med-PODiff (Coarse UQ) & 0.926 & 0.711 & 0.534 \\
\hline
PixelDiff & 0.892 & 0.650 & 0.576 \\
VAE-LDM & 0.855 & 0.564 & 0.560 \\
DiT & 0.903 & 0.681 & 0.587 \\
Full-field POD & 0.894 & 0.676 & 0.581  \\
\hline
\end{tabular}
\caption{Detection of high total-error pixels on the full test set
($N=2{,}130$ slices). Metrics are computed within the brain foreground
mask. Positives are the top $10\%$ of foreground pixels by absolute
reconstruction error in each image.}
\label{tab:hallucination_detection}
\end{table}
\subsection{Scale-Selective Uncertainty and Hallucination Localization}
\label{sec:coarse_fine_results}
\begin{figure*}[t]
\centering
\includegraphics[width=\textwidth]{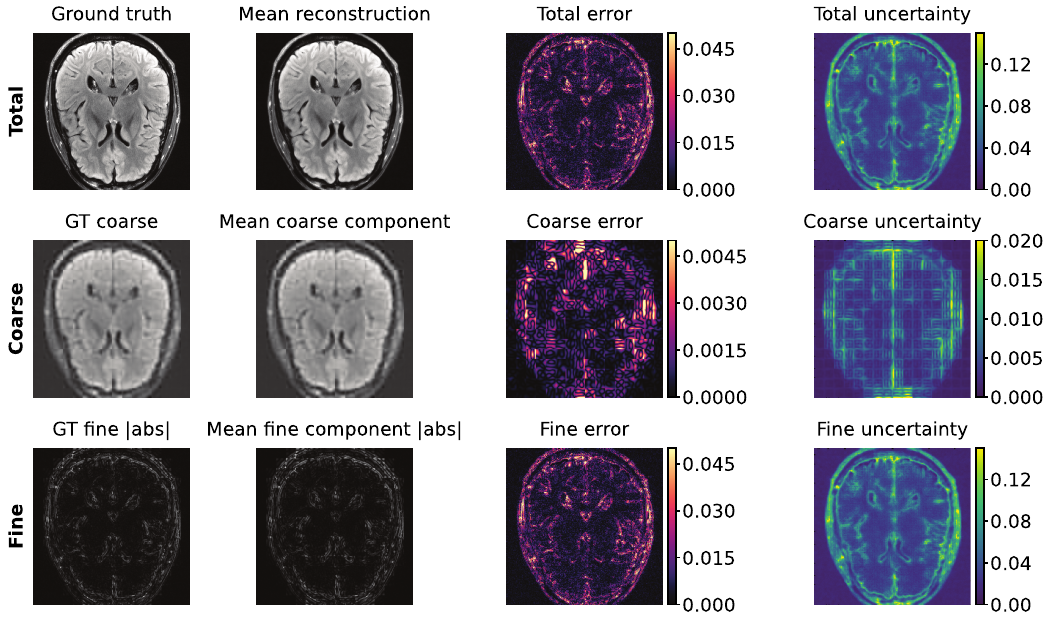}
\caption{Representative coarse--fine decomposition. Rows show total,
coarse, and fine components; columns show the target, predictive mean,
absolute component error, and uncertainty. Coarse and fine components
use modes $1,\ldots,k_c$ and $k_c+1,\ldots,K$, respectively, with
$k_c=10$. Signed fine components are shown in absolute value; the
coarse-panel grid reflects the non-overlapping $16\times16$ patches.}
\label{fig:decomposition}
\end{figure*}

Scale selectivity is evaluated on a fixed random subset of $N=200$
test images through a double-dissociation test: coarse uncertainty
should preferentially predict coarse error, and fine uncertainty should
preferentially predict fine error. Figure~\ref{fig:decomposition}
illustrates the separation: low-order modes recover broad anatomical and
intensity structure, whereas the remaining retained modes concentrate
around boundaries and localized detail, with their different magnitudes
reflected by the colorbar ranges. For the scale-specific analyses, we use a fixed volume-stratified subset of 200 slices drawn across all 186 test volumes. Each volume contributes at least one slice, and no volume contributes more than two slices.

Table~\ref{tab:decomposition} confirms the double dissociation, with
same-scale minus cross-scale AUROC gaps of $0.140$ for coarse and $0.094$
for fine uncertainty; AUPRC and Pearson correlation show the same
separation. The dominance-weighted $H_p$ retains strong fine-error
localization (AUROC $0.906$, Pearson $r=0.754$).
\begin{table}[t]
\centering
\small
\setlength{\tabcolsep}{2.0pt}
\renewcommand{\arraystretch}{0.96}
\begin{tabular}{@{}lcccc@{}}
\hline
Score $\rightarrow$ target
& Pearson $r$
& AUROC$\uparrow$
& AUPRC$\uparrow$
& P@10$\uparrow$ \\
\hline
$U_{\mathrm{tot}}\rightarrow e_{\mathrm{tot}}$
& 0.900 & 0.927 & 0.747 & 0.589 \\
$U_c\rightarrow e_c$
& 0.757 & 0.901 & 0.708 & 0.527 \\
$U_f\rightarrow e_f$
& 0.857 & 0.926 & 0.745 & 0.583 \\
\hline
$U_c\rightarrow e_f$
& 0.396 & 0.761 & 0.332 & 0.367 \\
$U_f\rightarrow e_c$
& 0.425 & 0.832 & 0.335 & 0.373 \\
\hline
$H_p\rightarrow e_f$
& 0.754 & 0.906 & 0.740 & 0.570 \\
\hline
\end{tabular}
\caption{Scale-specific uncertainty--error alignment on a fixed random
subset of $N=200$ test images. Here, $U_{\mathrm{tot}}$, $U_c$, and
$U_f$ denote total, coarse, and fine uncertainty, while
$e_{\mathrm{tot}}$, $e_c$, and $e_f$ denote the corresponding errors.
Metrics are computed within the brain foreground. The first block
contains matched-scale pairs and the second contains cross-scale
controls. The final row evaluates $H_p$ against fine-scale error. Per-image SE ranges from
$0.006$ to $0.009$.}
\label{tab:decomposition}
\end{table}
\paragraph{Sensitivity to the coarse--fine cutoff and POD truncation.}

For $k_c\in\{2,5,10,20,40,80\}$, the same-scale minus cross-scale
AUROC gap remains positive: $0.137$--$0.140$ for coarse uncertainty,
$0.086$--$0.096$ for fine uncertainty, and $0.163$--$0.171$ for
$H_p$ when comparing fine- and coarse-error localization. Thus, the
coarse--fine separation is stable and not specific to the default
$k_c=10$.

The retained $K=128$ modes capture $99.062\%$ of the training energy,
leaving $0.938\%$ in the discarded tail. On the fixed $N=200$ subset,
the discarded component has mean MAE $0.00155$, versus $0.04115$ for
retained fine error, giving tail-to-fine ratios of $0.0381$ per image
and $0.0376$ from aggregate means. Hence, the discarded tail is only
about $3.8\%$ of the retained fine-error magnitude, indicating that the
observed scale selectivity arises primarily within the retained POD
space rather than from omitted high-order content.
\paragraph{Controlled localization and degradation robustness.}
Because clinical hallucination labels are unavailable, we insert five
small edge- or blob-like high-frequency structures per slice before
degradation and evaluate localization on their known support.
Among reference-free scores, $H_p$ performs best
(Table~\ref{tab:additional_robustness}), achieving AUROC $0.926$,
AUPRC $0.711$, and P@10 $0.735$. Its performance is comparable to
reference-based sFRC (AUROC $0.928$, AUPRC $0.701$, P@10 $0.752$),
despite requiring no high-resolution target at inference. At
$\pi_{\mathrm{inj}}=0.20$, its AUPRC is $2.37\times$ the random
baseline. This experiment evaluates sensitivity to unresolved fine
detail rather than literal hallucination ground truth.

\begin{figure*}[t]
\centering
\includegraphics[width=0.9\textwidth]{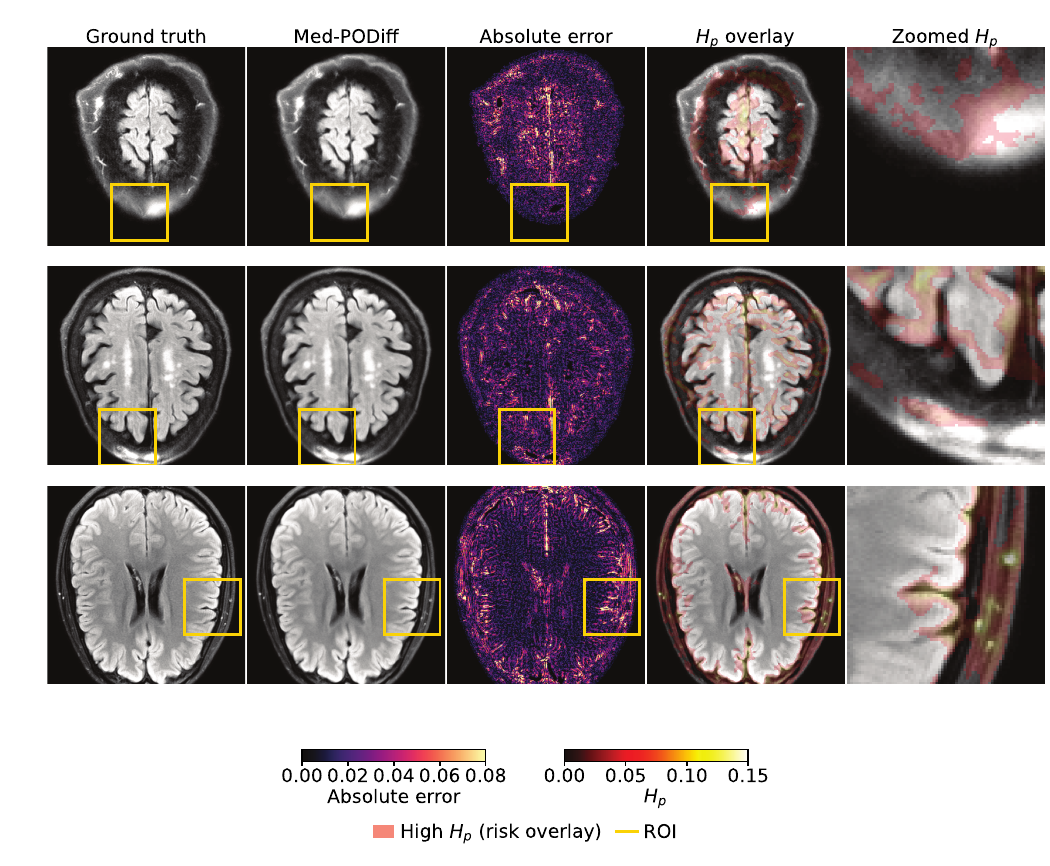}
\caption{Localization of fine-scale reconstruction risk in three
representative test cases. Columns show the high-resolution target,
Med-PODiff predictive mean, absolute reconstruction error, $H_p$
overlay, and an enlarged view of the selected region. Yellow boxes
denote the regions of interest. For visualization, the overlay retains
pixels above the per-image $70$th percentile of $H_p$; quantitative
metrics use the continuous score.}
\label{fig:hp_localization}
\end{figure*}

Under retrospective $4\times$ Cartesian $k$-space undersampling,
total uncertainty achieves total-error AUROC $0.934$, while $H_p$
retains fine-error AUROC $0.882$. The same-scale minus cross-scale
AUROC gaps remain positive for coarse ($0.115$) and fine ($0.056$)
uncertainty, showing that the decomposition is not specific to
bicubic degradation.
\begin{table}[t]
\centering
\small
\setlength{\tabcolsep}{2.5pt}
\begin{tabular}{lccc}
\hline
\multicolumn{4}{l}{\textit{(a) Injected-detail localization}}\\
Score & AUROC$\uparrow$ & AUPRC$\uparrow$ & P@10$\uparrow$\\
\hline
Total UQ & 0.877 & 0.668 & 0.615\\
Fine UQ  & 0.901 & 0.691 & 0.701\\
$H_p$    & 0.926 & \textbf{0.711} & 0.735\\
sFRC (HR-ref.) & \textbf{0.928} & 0.701 & \textbf{0.752}\\
\hline
\multicolumn{4}{l}{\textit{(b) Degradation robustness}}\\
\multicolumn{2}{l}{Metric} & Bicubic & $k$-space\\
\hline
\multicolumn{2}{l}{PSNR$\uparrow$} & 26.720 & 25.715\\
\multicolumn{2}{l}{SSIM$\uparrow$} & 0.9120 & 0.8990\\
\multicolumn{2}{l}{MACE$_4\downarrow$} & 0.0107 & 0.0196\\
\multicolumn{2}{l}{$U_{\mathrm{tot}}\!\rightarrow e_{\mathrm{tot}}$
AUROC$\uparrow$} & 0.927 & 0.934\\
\multicolumn{2}{l}{$H_p\!\rightarrow e_f$ AUROC$\uparrow$}
& 0.906 & 0.882\\
\hline
\end{tabular}
\caption{Controlled localization and degradation robustness on the
same fixed $N=200$ test images. Panel (a) uses injected-feature support
as the positive set ($\pi_{\mathrm{inj}}=0.20$). sFRC compares the
predictive mean with the high-resolution target, whereas the uncertainty
scores do not access that target at inference. Panel (b) compares
bicubic and retrospective Cartesian $k$-space degradation. Per-image SE ranges from
$0.005$ to $0.006$ in panel (a).}
\label{tab:additional_robustness}
\end{table}
\paragraph{Null-space hallucination localization.}
To distinguish generic reconstruction error from erroneous content
unsupported by the measurements, we use the retrospective Cartesian
operator $A=MF$ and define
$P_{\mathrm{range}}=A^{H}(AA^{H})^{-1}A$ and
$P_{\mathrm{null}}=I-P_{\mathrm{range}}$, which for Cartesian sampling
project onto the acquired and unacquired $k$-space coefficients,
respectively. We decompose the predictive-mean error into
$h=|P_{\mathrm{null}}(\bar{x}-x^{\mathrm{HR}})|$ and
$g=|P_{\mathrm{range}}(\bar{x}-x^{\mathrm{HR}})|$, where $h$ measures
erroneous content in the measurement null space and therefore provides a
direct proxy for data-unsupported hallucination.
\begin{table}[t]
\centering
\small
\setlength{\tabcolsep}{5pt}
\begin{tabular}{lccc}
\hline
Score & $h$ AUROC $\uparrow$ & $g$ AUROC $\uparrow$
& Gap $\uparrow$ \\
\hline
$H_p$       & \textbf{0.9186} & 0.8526 & \textbf{0.0660} \\
$U_{\mathrm{tot}}$ & 0.8826 & 0.8799 & 0.0027 \\
$U_f$       & 0.8789 & 0.8738 & 0.0051 \\
\hline
\end{tabular}
\caption{Null- versus range-space error localization. The gap
is AUROC$(h)$ minus AUROC$(g)$. Per-image SE ranges from
$0.005$ to $0.006$.}
\label{tab:nullspace}
\end{table}
As shown in Table~\ref{tab:nullspace}, $H_p$ preferentially localizes
null-space error, with a selectivity gap of $0.0660$, whereas total and
fine uncertainty are nearly indifferent between the two subspaces (gaps
of $0.0027$ and $0.0051$). Thus fine uncertainty alone behaves mainly as
a generic error indicator, while the dominance weighting in $H_p$
selectively emphasizes erroneous detail unsupported by the acquired
measurements. This provides the strongest evidence that $H_p$ acts as a
hallucination-risk detector rather than merely identifying high-error
regions. Figure~\ref{fig:hp_localization} illustrates the spatial
correspondence between $H_p$ and localized reconstruction error.

\subsection{Downstream Segmentation Consistency}
\label{sec:segmentation}

Aggregate image fidelity does not guarantee preservation of the structures used by downstream algorithms. We therefore compare foreground brain segmentations obtained from each reconstruction with
those obtained from the corresponding high-resolution target. Table~\ref{tab:segmentation} reports agreement over the full test set.

Med-PODiff achieves the best observed mean on all four segmentation-consistency metrics. Relative to bicubic interpolation, it increases Dice from $0.450$ to $0.775$ and IoU from $0.291$ to $0.632$, while reducing HD95 by $4.1\times$ and volumetric error by $9.5\times$. Relative to DiT, Dice and IoU increase by $0.016$ and $0.021$, respectively; HD95 decreases by $1.1$ voxels, and volumetric error decreases from $0.030$ to $0.021$.
\begin{table}[t]
\centering
\small
\setlength{\tabcolsep}{1.5pt}
\renewcommand{\arraystretch}{0.97}
\begin{tabular}{@{}lcccc@{}}
\hline
Method
& Dice$\uparrow$
& IoU$\uparrow$
& HD95$\downarrow$
& VolErr$\downarrow$ \\
\hline
Bicubic
& 0.450 (.027)
& 0.291 (.022)
& 53.6 (7.0)
& 0.200 (.056) \\
PixelDiff
& 0.742 (.035)
& 0.589 (.032)
& 16.2 (2.2)
& 0.033 (.022) \\
VAE-LDM
& 0.705 (.027)
& 0.543 (.021)
& 20.2 (2.0)
& 0.041 (.033) \\
DiT
& 0.759 (.040)
& 0.611 (.042)
& 14.2 (2.0)
& 0.030 (.028) \\
\textbf{Med-PODiff}
& \textbf{0.775 (.036)}
& \textbf{0.632 (.031)}
& \textbf{13.1 (1.6)}
& \textbf{0.021 (.029)} \\
\hline
\end{tabular}
\caption{Downstream foreground-segmentation consistency on the full test split
($186$ volumes; $2{,}130$ slices). SynthSeg foreground masks from the
high-resolution images serve as references. HD95 is reported in voxel
units, and VolErr is the absolute relative foreground-volume error.
Entries are mean (SD) across volumes.}
\label{tab:segmentation}
\end{table}

\paragraph{Localization of downstream segmentation disagreement.}

On a fixed random subset of $N=200$ test images, $H_p$ provides the
strongest segmentation-disagreement localization. For $H_p$, total
uncertainty, and fine uncertainty, respectively, Pearson $r$ is
$0.8291$, $0.8106$, and $0.7836$; AUROC is $0.8011$, $0.7954$, and
$0.7800$; AUPRC is $0.7257$, $0.7029$, and $0.7024$; and P@10 is
$0.6385$, $0.6059$, and $0.5897$. At disagreement prevalence
$\pi_D=0.3155$, the AUPRC of $H_p$ is $2.30\times$ the random
baseline, while its P@10 indicates that $63.85\%$ of pixels in the
highest-risk decile are disagreement pixels. Although its AUROC gain
over total uncertainty is modest, the larger AUPRC and P@10 gains show
that $H_p$ concentrates disagreement more effectively among high-risk
predictions.

This complements Table~\ref{tab:decomposition}: fine uncertainty
captures generic fine-scale error, whereas $H_p$ emphasizes regions
where fine uncertainty is both large and dominant, yielding a more
selective indicator of downstream structural disagreement.
%
%
\paragraph{Statistical significance.}
Scale-selectivity and localization metrics are computed per image for all test samples. We report $95\%$ confidence intervals from bootstrap
resampling over images ($B=10{,}000$) and test key gaps with a two-sided
paired Wilcoxon signed-rank test over the $200$ per-image differences. The
coarse and fine double-dissociation gaps are $0.140$ $[0.118,\,0.156]$ and
$0.094$ $[0.083,\,0.109]$, both $p<10^{-6}$. The null-space selectivity of
$H_p$ is $0.066$ $[0.031,\,0.099]$, $p<10^{-3}$, whereas the gaps for
$U_\mathrm{tot}$ and $U_f$ are not significant ($p=0.45$, $0.29$).
Improvements over DiT are significant for RMSE, MAE, and PSNR
($p<10^{-4}$); the SSIM margin is directional ($\Delta=0.0031$,
$p=0.059$).
\paragraph{Efficiency.}
Med-PODiff uses 40M parameters, 6.6\,GB peak training memory, and
0.091\,s per sample, compared with 45M, 9.1\,GB, and 0.186\,s for
DiT and 160M, 13.5\,GB, and 1.854\,s for PixelDiff.

\section{Limitations}
Med-PODiff assumes locally low-rank structure; slower POD decay would
require more modes, while its fixed training-derived basis motivates
future adaptive or online basis learning. POD ordering reflects
explained variance rather than spatial frequency, so the coarse--fine
interpretation remains empirical, although stable across mode cutoffs.
Discarded-mode uncertainty is not modeled, but its measured magnitude
is only $3.8\%$ of the retained fine-error magnitude. Moreover, $H_p$
is a ranking score rather than a calibrated clinical probability.
Evaluation remains limited to fastMRI, synthetic bicubic and
retrospective Cartesian degradations, and SynthSeg-derived rather than
manual labels; external validation across scanners, protocols,
contrasts, and pathologies is needed.

\bibliography{aaai2027}

\end{document}